\documentclass{article}
\usepackage{spconf,amsmath,graphicx,hyperref}
\usepackage{multirow}
\usepackage{algorithm}
\usepackage{algorithmic}
\usepackage{amssymb}
\usepackage{amsfonts}

\usepackage{cite}
\title{UniGeo: A Unified 3D Indoor Object Detection Framework Integrating Geometry-Aware Learning and Dynamic Channel Gating}
\name{Xing Yi$^{1}$, Jinyang Huang$^\star$$^{1}$, Feng-Qi Cui$^{2}$, Anyang Tong$^{1}$, Ruimin Wang$^{1}$, Liu Liu$^{1}$, Dan Guo$^{1}$ \thanks{$^\star$ Corresponding author: hjy@hfut.edu.cn} }

\address{$^{1.}$ School of Computer Science and Information Engineering, Hefei University of Technology, Hefei, China \\
      $^{2.}$ Institute of Advanced Technology, University of Science and Technology of China, Hefei, China}

\begin{document}
\ninept
\maketitle
\begin{abstract}
% The expanding applications of robotics and augmented reality in real-world scenarios have drawn significant research attention to point cloud-based 3D object detection technologies. 

The growing adoption of robotics and augmented reality in real-world applications has driven considerable research interest in 3D object detection based on point clouds.
While previous methods address unified training across multiple datasets, they fail to model geometric relationships in sparse point cloud scenes and ignore the feature distribution in significant areas, which ultimately restricts their performance. To deal with this issue, a unified 3D indoor detection framework, called UniGeo, is proposed. To model geometric relations in scenes, we first propose a geometry-aware learning module that establishes a learnable mapping from spatial relationships to feature weights, which enabes explicit geometric feature enhancement. Then, to further enhance point cloud feature representation, we propose a dynamic channel gating mechanism that leverages learnable channel-wise weighting. This mechanism adaptively optimizes features generated by the sparse 3D U-Net network, significantly enhancing key geometric information.
Extensive experiments on six different indoor scene datasets clearly validate the superior performance of our method.

\end{abstract}
\section{Introduction}
\label{sec:intro}

Indoor 3D object detection is a core task in scene understanding with applications in robotics, pose estimation\cite{Break} and AR/VR systems. Traditional methods~\cite{cagroup3d, votenet, sofw, cpdet3d, mlcvnet, detr3d, uni3detr, spgroup3d, groupfree, brnet, vdetr, tr3d, fcaf3d} struggle to generalize, relying heavily on specific datasets and failing to build a unified learning framework. This limitation stems from their inability to decouple geometric features from semantic information, leading models to overfit dataset-specific surface characteristics rather than capturing essential object attributes.

% To address these limitations, Kolodiazhnyi proposed UniDet3D \cite{unidet3d}, the first unified multi-dataset 3D detection framework enabling collaborative training across diverse 3D data sources. This approach breaks domain barriers inherent in single-dataset training and significantly improves generalization, laying a foundation for universal 3D detection. 
% %
% However, the sparse 3D U-Net backbone used in UniDet3D has notable weaknesses in geometric feature modeling and spatial representation. It separately processes coordinates and voxel features, causing geometric structure information to diminish in deeper layers and lacking an explicit spatial modeling mechanism. 
% % some works are proposed to address the limitation of UniDet3D, ...(介绍下别人如何解决的，但是这些方法存在哪些问题)
% Moreover, its inability to adapt to the uneven spatial distribution of point clouds hampers distinguishing complex regions (e.g., edges, corners) from flat areas. Fixed kernel sizes and receptive fields further prevent dynamic adjustment to local geometric complexity, limiting multi-scale feature extraction. Finally, feature aggregation overlooks spatial position contributions, diluting key geometric features—especially in indoor scenes with many low-information points like walls and floors. Collectively, these issues constrain sparse 3D U-Nets in complex scene understanding and detection tasks. 
To overcome these limitations, the State-of-the-art (SOTA) method proposed UniDet3D~\cite{unidet3d}, a unified multi-dataset 3D detection framework that allows collaborative training across multiple data sources, enhancing generalization and building a basis for generic 3D detection. However, this method adopted a sparse 3D U-Net backbone with significant shortcomings, i.e., it processed spatial coordinates and voxel characteristics independently, leading geometric information to a decline in deeper layers and lacking explicit spatial connection modeling. Additionally, this method also failed to adapt to non-uniform point cloud distributions, challenging to detect geometrically different areas like edges and corners.  These limitations jointly hinder the model's performance in complicated 3D scene interpretation tasks.

In this paper, we propose UniGeo, a unified 3D indoor detection framework that incorporates geometric-aware learning and a dynamic channel gating mechanism. Specifically,
% we introduce a multi-scale feature extraction strategy based on the sparse 3D U-Net backbone, which can enhance the backbone's ability to capture information ranging from local details to global structures. Secondly,
we introduce a geometric-aware learning module to model scene geometric relationships, which establishes a mapping from spatial scene geometric relationships to feature weights, allowing for explicit feature improvement based on 3D geometric structures. 
Firstly, we employ Euclidean distance to model the geometric topology of point clouds in the scene, and then assign differentiated weights to point cloud features using an exponential decay function based on the topological information, thereby enhancing focus on significant regions.
% to model spatial geometric relationships in the scene, we design a Geometric-aware Learning Module, which establishes a learnable mapping from spatial relationships to feature weights, enabling explicit feature enhancement based on 3D geometric structures.
% Furthermore, to address the backbone network's limited capability in adaptively modeling the spatial distribution of point clouds, we propose a Dynamic Channel Gating mechanism. Tailored to the characteristics of sparse 3D data, DCG dynamically modulates the feature response strength across different channels, effectively enhancing discriminative geometric information while suppressing interference from background noise.
Then, to further address the issue of the backbone network modeling point cloud's limited spatial distribution capability, we suggest a dynamic channel gating mechanism that adaptively learns the channel weight to help modulate the characteristic response strength on various channels, efficiently enhance the local characteristic information, and suppress the background noise information.
% To evaluate the effectiveness and generalization capability of the proposed UniGeo architecture, we conducted comprehensive experiments on six indoor scene datasets: ScanNet, S3DIS, MultiScan, 3RScan, ScanNet++, and ARKitScenes. The results demonstrate that our method achieves state-of-the-art performance, outperforming existing mainstream approaches in terms of both best and average performance on most datasets. Furthermore, through a series of ablation studies on model components, algorithm selection comparisons, and hyperparameter analysis, we thoroughly validate the superiority of our method.

To evaluate the effectiveness and generalization capability of UniGeo, we conduct detailed experiments on six indoor scene datasets: ScanNet~\cite{scannet}, S3DIS~\cite{s3dis}, MultiScan~\cite{multiscan}, 3RScan~\cite{3rscan}, ScanNet++~\cite{scannet++}, and ARKitScenes~\cite{arkitscenes}. The results demonstrate that our method achieves state-of-the-art performance, outperforming existing mainstream approaches in terms of both best and average performance on most datasets. Furthermore, through a series of ablation studies on model components, algorithm selection comparisons, and hyperparameter analysis, we thoroughly validate the superiority of UniGeo. 
The main contributions are as follows:
\begin{figure*}[t]
  \centering
  \includegraphics[width=0.98\linewidth]{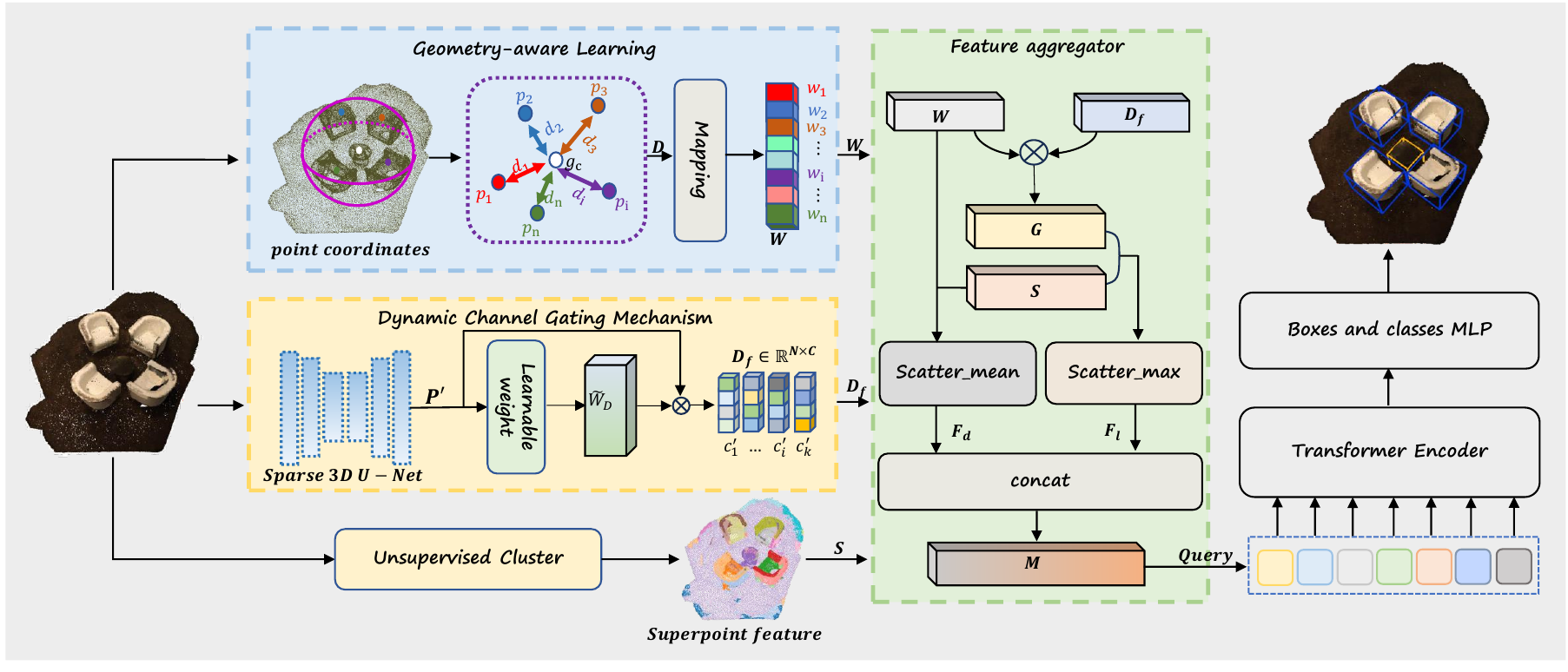}
  \vspace{-15pt}
  \caption{Overview of our method. UniGeo takes a point cloud as input and pass through a geometry-aware learning module and a dynamic channel gating mechanism to generate geometry weights and channel features. A feature aggregator combines these with superpoint features into a hybrid representation, which then serves as input queries to a transformer encoder. Finally, a box MLP and a class MLP predict the 3D bounding boxes from the transformer encoder outputs.}
    \vspace{-10pt}
    \label{main}
\end{figure*}
\begin{itemize}
% \item We propose a unified 3D indoor scene detection framework incorporating geometric awareness and an adaptive gating mechanism. This framework effectively addresses the limitations of previous methods in modeling geometric features and spatial relationships, thereby enhancing the decoupling capability between scene features and semantic information.

\item To model geometric relationships in sparse point cloud scenes and achieve high detection accuracy, a unified 3D indoor detection framework, UniGeo, is proposed. By introducing a geometric-aware learning and a dynamic channel gating mechanism to establish a learnable mapping from spatial relationships to feature weights, we effectively handle non-uniformly distributed scene point clouds and achieve explicit feature enhancement based on 3D scene geometric structures.

\item Extensive evaluation experiments across various metrics on six indoor scene detection datasets consistently demonstrate the superior performance of our proposed method.
\end{itemize}

\section{Method}
\label{sec:Method}
\subsection{Approach Overview}
The overall architecture of our method is depicted in Fig.~\ref{main}. The input point cloud typically consists of N points, which can be represented as $\mathcal{P} = \left ({ p_i|i = 1,...,N}\right ) \in \mathbb{R}^{N\times6}$. Each point has coordinates $x$, $y$, $z$ and colors $r$, $g$, $b$. Following previous methods, 
% we first extract the point features $\mathcal {P}' = \left ({ {p_i}'|i = 1,...,N}\right )\in \mathbb{R} ^{N\times C} $ using a sparse 3D UNet network. 
% Next, these point features are processed through dynamic channel gating Mechanism module to obtain geometric learning features and channel-enhanced features, 
% Subsequently, we aggregate these two features with the superpoint features ${S}= \left ({ s_i|i = 1,...,N}\right ) \in \mathbb{R} ^{M\times C} $ obtained from unsupervised learning clustering of the point cloud data through a feature aggregator module to generate proposals.
we first take the point cloud $\mathcal{P}$ as the input of the geometry-aware learning module and the dynamic channel gating mechanism and obtain their spatial weighting coefficients $\mathcal{W} = \{w_i|i = 1,...,N\}\in \mathbb{R}^{N\times 1}$ and channel feature $\mathcal D_f = \{d_{f}^{i}|i = 1,...,N\}\in \mathbb{R} ^{N\times C}$, respectively. Meanwhile, we obtain the superpoint feature ${S}= \left ({ s_i|i = 1,...,N}\right ) \in \mathbb{R} ^{M\times C} $ from the point cloud by unsupervised clustering. Then, we aggregate these features through a feature aggregator and obtain a hybrid representation $\mathcal M$.
Finally, the hybrid representation $\mathcal M$ serve as queries for the transformer encoder. 
After processing through the transformer encoder, the output is fed into boxes and classes MLP  to predict the detection boxes and classes results.

\subsection{Geometry-aware Learning}
% The traditional sparse 3D UNet architecture exhibits several limitations when processing point cloud data. Firstly, conventional sparse 3D UNets typically process spatial coordinates and features separately, failing to explicitly model geometric relationships between points. Secondly, they lack an adaptive feature weighting mechanism based on spatial distribution, which is crucial for handling non-uniform point densities commonly encountered in real-world 3D scenes.

% We introduce a geometry-aware learning module that addresses the inherent limitations of traditional sparse 3D U-Nets in geometric feature representation. By establishing an explicit spatial relationship modeling mechanism, this module achieves differentiated feature processing for different regions of the point cloud through a nonlinear mapping between spatial distances and feature importance. This approach effectively resolves the fundamental shortcomings of conventional sparse 3D U-Nets in geometric feature representation.
 A Geometry-Aware Learning(GAL) module that models spatial geometric structures is first proposed and to establish a weight mapping between topological information and point cloud data. By adaptively weighting scene features, the module enhances the model's focus on significant regions.
Specifically, we take the 3D coordinates $\mathcal{P} = \{ {p_i}=({x_i},{y_i},{z_i})|i = 1,...,N\}\in \mathbb{R}^{N\times3}$ as input of the geometry-aware learning module, and we further compute the geometric centroid $g_c$ as the reference point to model geometric structures. In particular, $g_c$  can be calculated as:
\begin{equation}
g_c = (\bar{x}, \bar{y}, \bar{z}) = \left( \frac{1}{N} \sum_{i=1}^N {x_i}, \frac{1}{N} \sum_{i=1}^N {y_i}, \frac{1}{N} \sum_{i=1}^N {z_i} \right).
\end{equation}

To model the importance and geometric relationships of different regions in scenes, we compute the Euclidean distance from each sampled point $\mathcal{P}' = \{ {p_i}'|i = 1,...,N\}\in \mathbb{R}^{N\times3}$ to the geometric centroid $g_c$, thereby learning geometric feature representations. The Euclidean distance $\mathcal{D} = \{ d_i|i = 1,...,N\}\in \mathbb{R}^{N\times 1}$ can be represented as:
\begin{equation}
d_i = \| {p_i} - g_c \|_2 = \sqrt{({x_i} - \bar{x})^2 + ({y_i} - \bar{y})^2 + ({z_i} - \bar{z})^2}.  
\end{equation}
To preserve the consistency of the point cloud topological structure, we employ min-max scaling normalize the Euclidean distance representation $\mathcal{D}$, thus obtaining the normalized spatial distance feature $\mathcal{\tilde{D}} = \{ \tilde{d}_{i}|i = 1,...,N\} \in \mathbb{R}^{N\times 1}$.

% To achieve adaptive characterization of global geometric structures, we construct a scale-aware feature modulation space through distance-weighted mapping. By applying an exponentially decaying function to the normalized features $\mathcal{\tilde{D}}$, we assign spatially-adaptive weights to obtain spatial weighting coefficients $\mathcal{W} = \{w_i|i = 1,...,N\}\in \mathbb{R}^{N\times 1}$ are calculated as follows:
To enhance the global geometric structure of the scene, we employ an exponential decay function to process the normalized feature $\mathcal{\tilde{D}}$. This constructs a mapping between the spatial topological structure and the scene point cloud data, and ultimately obtains the spatial weighting coefficients $\mathcal{W} = {w_i \mid i = 1,\ldots,N} \in \mathbb{R}^{N \times 1}$ for each sample point. The weighting coefficients $\mathcal{W}$ are calculated as:
\begin{equation}
w_i = \exp(-\alpha \cdot \tilde{d}_i),
\end{equation}
where the hyperparameter $\alpha$ denotes the exponential decay coefficient, and we set this hyperparameter to $\alpha=2$. 

 \begin{table*}[t]
\centering
\resizebox{1.0\linewidth}{!}{
\begin{tabular}{cccccccccccccc}
\hline
\multirow{2}{*}{Method} &\multirow{2}{*}{Venue} &\multicolumn{2}{c}{ScanNet} &\multicolumn{2}{c}{ARKitScenes} &\multicolumn{2}{c}{S3DIS} &\multicolumn{2}{c}{MultiScan} &\multicolumn{2}{c}{3RScan} &\multicolumn{2}{c}{ScanNet++} \\
 &&$mAP_{25}$ &$mAP_{50}$ &$mAP_{25}$ &$mAP_{50}$ &$mAP_{25}$ &$mAP_{50}$ &$mAP_{25}$ &$mAP_{50}$ &$mAP_{25}$ &$mAP_{50}$ &$mAP_{25}$ &$mAP_{50}$ \\\hline
\multicolumn{14}{c}{Best Results} \\
\hline
FCAF3D\cite{fcaf3d} &ECCV'2022 &71.5 &57.3 & & &66.7 &45.9 &53.8 &40.7 &60.1 &42.6 &22.3 &11.4 \\
TR3D\cite{tr3d} &ICIP'2023 &72.9 &59.3 & & &74.5 &51.7 &56.7 &42.3 &62.3 &45.4 &26.2 &14.5 \\
SPGroup3D\cite{spgroup3d} & AAAI'2024 & 74.3 & 59.6 &  & &69.2 &47.2 & & & & & & \\
V-DETR\cite{vdetr}&ICLR'2024 &77.4 &65.0 & & & & & & & & & & \\
SOFW\cite{sofw} &TMM'2025 &70.9 &52.3 & & & & & & & & & & \\
UniDet3D\cite{unidet3d} &AAAI'2025 &77.5 &\textbf{66.1} &\textbf{61.3} &47.1 &75.2 &60.8 &64.2 &51.6 &64.7 &48.6 &\textbf{26.4} &17.2 \\
\textbf{UniGeo (Ours)} &-&\textbf{77.7} &65.6 &60.0 &\textbf{47.7} &\textbf{80.5} &\textbf{71.8} &\textbf{69.6} &\textbf{56.3} &\textbf{68.1} &\textbf{56.1} &25.6 &\textbf{19.1} \\\hline
\multicolumn{14}{c}{Average across 25 trials} \\\hline
FCAF3D\cite{fcaf3d} &ECCV'2022 &70.7 &56.0 & & &64.9 &43.8 &52.5 &39.2 &59.6 &40.4 &21.4 &11.0 \\
TR3D\cite{tr3d} &ICIP'2023 &72.0 &57.4 & & &72.1 &47.6 &55.0 &41.2 &61.5 &44.2 &24.3 &13.9 \\
SPGroup3D\cite{spgroup3d} &AAAI'2024 &73.5 &58.3 & & &67.7 &43.6 & & & & & & \\
V-DETR\cite{vdetr} &ICLR'2024 &76.8 &64.5 & & & & & & & & & & \\
% DiffVote &TCSVT'2024 &70.9 &55.2 & & & & & & & & & & \\
UniDet3D\cite{unidet3d} & AAAI'2025 &\textbf{77.1} &\textbf{65.2} &\textbf{60.2} &46.0 &73.3 &57.9 &62.4 &50.8 & 62.1 & 45.6 & 24.4 & 16.3 \\
\textbf{UniGeo (Ours)} &-&76.8 &64.5 &59.3 &\textbf{46.8} &\textbf{77.9} &\textbf{66.3} &\textbf{65.9} &\textbf{53.5} &\textbf{64.2} &\textbf{49.7} &\textbf{24.6} &\textbf{17.8} \\\hline
\end{tabular}}
\vspace{-10pt}
 \caption{ Comparison with state-of-the-arts methods on 6 indoor scenes datasets, UniGeo demonstrates superior performance. Best results indicate the model's best evaluation results, Average across 25 trials represents the average evaluation score obtained by training 5 times and test each trained model 5 times independently.}
\label{tab:main}
\vspace{-18pt}
\end{table*}
\subsection{Dynamic Channel Gating Mechanism}
% In 3D indoor scene object detection, the features extracted by sparse 3D U-Net exhibit two critical issues: (1) Different channels contribute unequally to the final detection task—for instance, structural feature channels (e.g., geometric edges and object surface curvature) exhibit higher discriminative power than those from homogeneous regions; (2) A large number of background voxels (e.g., walls and floors) may dominate certain channels with irrelevant features. Traditional attention mechanisms suffer from prohibitively high computational costs on sparse 3D data and struggle to adapt to dynamically changing sparsity patterns.

We propose a dynamic channel gating mechanism that employs learnable channel weights to adaptively optimize point cloud features generated by the sparse 3D U-Net network. This approach enhances key geometric information while effectively suppressing background noise and irrelevant features. Specifically, we first define a trainable weight vector $\mathcal{\tilde{W}_D}=\left [ w_{1}^{d}, w_{2}^{d},...,w_{c}^{d} \right ] \in \mathbb{R} ^{C}$ and initialize as $w_{c}^{d} = 0.1$. Then, the learned weight parameters are transformed into channel gating coefficients through a sigmoid function, where these coefficients represent the importance of each channel. To enhance geometric  feature channels (e.g., edges) and suppress irrelevant background information, we combines the learned gating coefficients with the point features $\mathcal{P}' = \{ {p_i}'|i = 1,...,N\}\in \mathbb{R}^{N\times C}$ generated for each point by the sparse 3D U-Net. The channel feature $\mathcal D_f = \{d_{f}^{i}|i = 1,...,N\}\in \mathbb{R} ^{N\times C}$ can be calculated as:
\begin{equation}
 \mathcal D_f = \left [w_{i}^{d} \odot {p_i}'|i=1, 2,..., N \right ]\in \mathbb{R} ^{N\times C},
\end{equation}
 % Where $C$ represent the size of channel. The advantages of our design lie in its scalability to large-scale sparse point cloud data, where the gating operation preserves the sparse structure of input features and avoids dense matrix computations. Moreover, it effectively enhances critical geometric features while suppressing background noise.
 where $C$ is the number of channel. 
 % The benefits of  dynamic channel gating lay in its adaptability to large-scale sparse point cloud data, where the gating operation keeps the sparse structure of input features and avoids dense matrix calculations. Moreover, it efficiently enhance the key geometric features and minimizing background noise.

 \subsection{ Feature Aggregation}

 We use unsupervised clustering to convert the point cloud $\mathcal {P}= \left [ p_i|i = 1,...,N\right ]\in \mathbb{R} ^{N\times 6}$ into superpoint~\cite{superpoint} features $ S \in \mathbb{R} ^{M\times C}$ according to the principles of UniDet3D\cite{unidet3d}. Then, a hybrid feature fusion approach is proposed for the feature aggregation step that combines the benefits of geometry-aware feature $\mathcal G$ and dynamic channel gating features $\mathcal D_f$ to efficiently incorporate multi-dimensional data. Particularly, we obtain the geometry-aware features $\mathcal G$ through element-wise multiplication between the weight parameters $\mathcal{W} = \{w_i|i = 1,...,N\}\in \mathbb{R} ^{N\times 1}$ and the dynamic channel gating features $\mathcal D_f = \{d_{f}^{i}|i = 1,...,N\}\in \mathbb{R} ^{N\times C}$, achieving spatially adaptive feature recalibration. Geometry-aware features $\mathcal{G}= \left [ g_i|i = 1,...,N\right ]\in \mathbb{R} ^{N\times C}$ can be calculated as:
\begin{equation}
g_i = (w_i \odot d_{f}^{i}|i=1,2,...N).
\end{equation}

Then, we aggregate the Geometry-aware features $\mathcal G$ with the superpoints feature $S= \left [ s_i|i = 1,...,N\right ] \in \mathbb{R} ^{M\times C}$ by using a scattered mean to obtain the global geometric feature representation $F_d$, which can be expressed as:
\begin{equation}
 \mathcal F_d = scatter\_mean\left [s_i, g_i|i=1,2,...,M \right])\in \mathbb{R} ^{M\times C }.
\end{equation}

 Meanwhile, we aggregate the dynamic channel gating features $\mathcal D_f = \{d_{f}^{i}|i = 1,...,N\}\in \mathbb{R} ^{N\times C}$ with the superpoints feature $S= \left [s_i|i = 1,...,N\right ] \in \mathbb{R} ^{M\times C}$ using a scattered max to obtain most discriminative local feature $F_l$, which is denoted by:
 \begin{equation}
 \mathcal F_l = scatter\_max\left [s_i, d_i|i=1,2,...,M \right])\in \mathbb{R} ^{M \times C}.
\end{equation}

Subsequently, The hybrid feature representation$\mathcal M$ is obtained by concatenating the global geometric feature  $\mathcal F_d$ with the local feature $\mathcal F_l$. $\mathcal M$ remains sensitive to the overall geometric layout while preserving critical local details. $\mathcal M$ can be calculated as:
\begin{equation}
 \mathcal M = concat\left [F_d, F_l\right])\in \mathbb{R} ^{M \times C}.
\end{equation}
 
\subsection{Transformer Encoder and Loss}
Following the stage of feature aggregation, the backbone features are processed by a transformer encoder network. This network takes $\mathcal M$ queries as input and outputs $\mathcal M$ object proposal features. Besides, the transformer architecture relies solely on self-attention mechanisms among the input queries. For the matching strategy and training objective, we adopt the same approach as used in UniDet3D~\cite{unidet3d}. The total loss $\mathcal L_{total}$ can be defined as:
\begin{equation}
\mathcal L_{total}= \beta L_{cls} + L_{reg},
\end{equation}
% Classification loss  $\mathcal L_{cls}$ are penalized using a cross-entropy loss. For each matched proposal-ground truth pair, the regression loss $\mathcal L_{reg}$ is computed as a DIoU loss between the predicted and ground truth bounding boxes. $\beta$ is the weight hyperparameter of $\mathcal L_{cls}$, and we set the value to 0.5.
where the classification loss $\mathcal L_{cls}$ is computed with a cross-entropy loss. For each matched proposal-ground truth pair, the regression loss $\mathcal L_{reg}$ is computed as a DIoU loss between the predicted and ground truth bounding boxes. $\beta$ is the weight hyperparameter for $\mathcal L_{cls}$, and we set it to 0.5 based on experience.

\section{Experimental}
\subsection{Experimental Setup}
\textbf{Datasets and metrics:} We conduct our experiments on six real-world indoor detection datasets, including ScanNet\cite{scannet}, S3DIS\cite{s3dis}, ARKitScenes~\cite{arkitscenes}, ScanNet++\cite{scannet++}, 3RScan\cite{3rscan}, and MultiScan\cite{multiscan}. The specific data categories and detailed information are provided in Table~\ref{tab:dataset}. We use mean average precision (mAP) with thresholds of 0.25 and 0.5 as the evaluation metrics. To obtain statistically significant results, we run training 5 times and test each trained model 5 times independently. Furthermore, we report the best and average metrics across 5 × 5 trials to ensure consistency with the experimental standards of other methods.

\begin{table}[h]
\centering
\resizebox{0.85\linewidth}{!}{
\begin{tabular}{cccc}
\hline
Dataset Name & \#Train Scenes & \#Val Scenes & \#Classes \\
\hline
ScanNet      & 1201           & 312          & 18        \\
ARKitScenes  & 4493           & 549          & 17        \\
S3DIS        & 204            & 68           & 5         \\
MultiScan    & 174            & 42           & 17        \\
3RScan       & 385            & 47           & 18        \\
ScanNet++    & 230            & 50           & 84        \\
\hline
Overall      & 6687           & 1068         & 99        \\
\hline
\end{tabular}}
\vspace{-10pt}
\caption{Quantitative statistics of indoor datasets in our mixture.}
\label{tab:dataset}
\vspace{-10pt}
\end{table}

\textbf{Parameter settings:} We implement our method in the MMDetection3D~\cite{mmdet3d} framework. The AdamW optimizer is used with an initial learning rate of 0.0002, a weight decay of 0.05, a batch size of 1, and a polynomial scheduler with a base of 0.9 for 1024 epochs. All experiments are conducted on four RTX 2028 Ti GPUs.
% Comparison Methods: We compared three baseline methods, namely FCAF3D, TR3D, and UniDet3D. FCAF3D uses fully convolutional networks for 3D object detection, eliminating the need for anchor points, which enhances the flexibility and accuracy of detection; TR3D is an extension of the FCAF3D method, primarily used for real-time 3D object detection in indoor scenes, thereby broadening the application range of FCAF3D. UniDet3D proposes a unified detection framework for indoor multi-dataset scenarios. Its advantage lies in not needing to train specific weight models based on the characteristics of different datasets, thus not being limited to specific scenarios, which in turn expands the range of applicable scenarios.
\subsection{Analysis of Results}
\begin{figure}[t]
  \centering
  \includegraphics[width=0.98\linewidth]{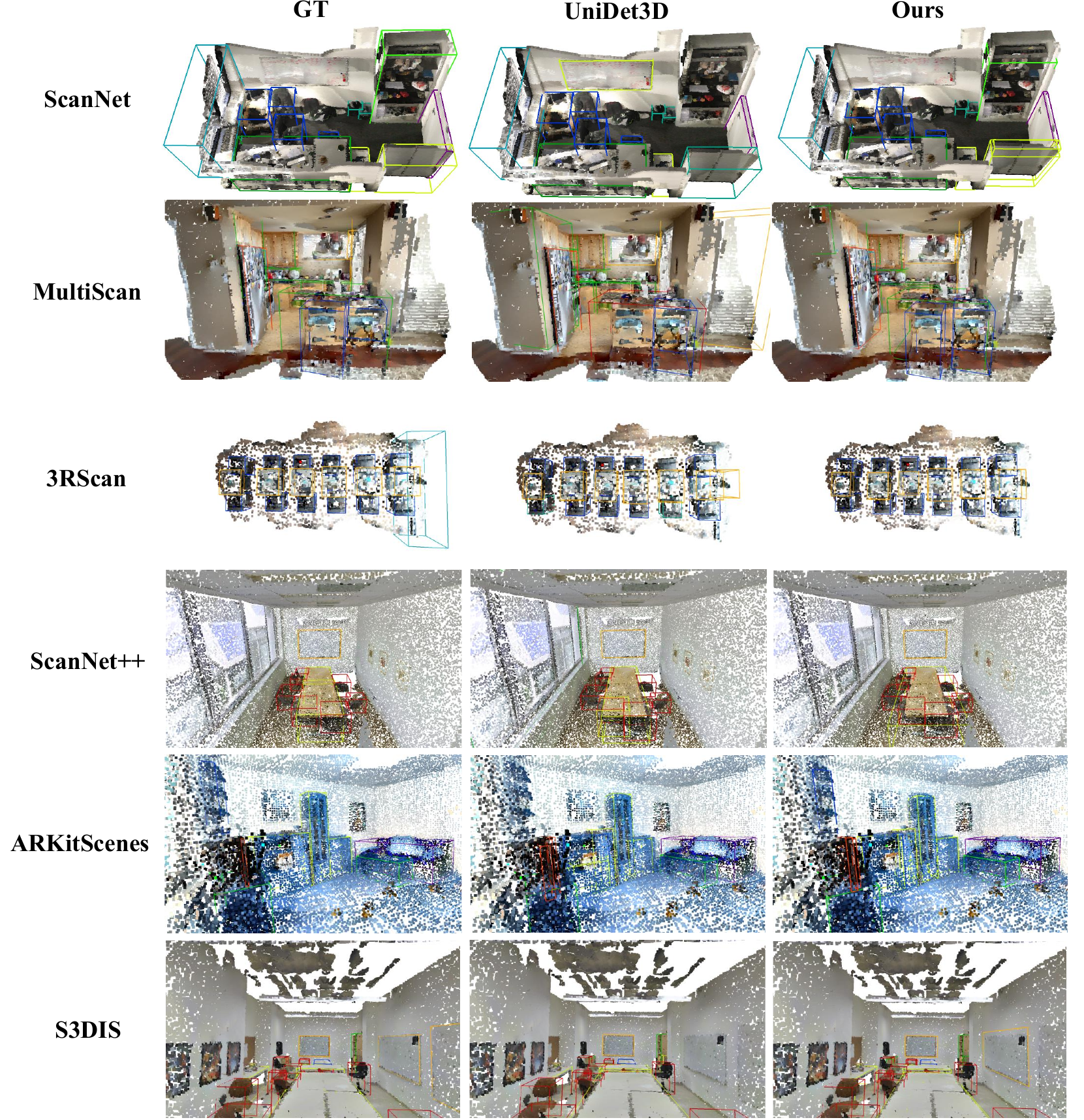}
   \vspace{-12pt}
  \caption{Qualitative comparison on six indoor scenes datasets.} 
     \vspace{-20pt}
    \label{fig:dataset}
\end{figure}

We compare UniGeo with the latest methods on six datasets: ScanNet~\cite{scannet}, S3DIS~\cite{s3dis}, MultiScan~\cite{multiscan}, 3RScan~\cite{3rscan}, and ScanNet++~\cite{scannet++}. The quantitative results are presented in Table~\ref{tab:main}, and qualitative results are shown in Fig.~\ref{fig:dataset}. In the best results, our method achieves state-of-the-art performance on most datasets. Particularly, our approach attains gains of \textbf{5.3\%} and \textbf{11\%} in $mAP_{25}$ and $mAP_{50}$ over UniDet3D~\cite{unidet3d} on the S3DIS~\cite{s3dis} dataset. 
% It also achieves gains of \textbf{5.4\%} and \textbf{4.7\%} on the MultiScan~\cite{multiscan} dataset and \textbf{3.4\%} and \textbf{8.2\%} on the 3RScan~\cite{3rscan} dataset, surpassing all previous methods. 
In the average across 25 trials setting, our method maintains strong performance and leads by \textbf{4.6\%} and \textbf{8.4\%} on the S3DIS~\cite{s3dis} dataset. 
Our method's superior performance stems from its enhanced focus on perceiving and learning critical scene features, coupled with its advanced modeling of scene geometry.
% \textbf{3.5\%} and \textbf{2.7\%} on the MultiScan~\cite{multiscan} dataset, \textbf{2.1\%} and \textbf{4.1\%} on the 3RScan~\cite{3rscan} dataset, and \textbf{0.2\%} and \textbf{1.5\%} on the ScanNet++~\cite{scannet++} dataset. 
Although our approach performs better on most datasets and metrics, significant domain differences among datasets limit information capture in ambiguous scenarios, leading to slightly lower accuracy on certain metrics.

\subsection{Ablation Study}
To evaluate the combined efficiency of the dynamic channel gating (DCG) mechanism and geometric-aware learning (GAL), we design an ablation experiment for the module. The experiment evaluates the following three configurations: (1) using only DCG, (2) using only GAL, (3) combining both DCG and GAL. The results are shown in Table~\ref{tab:GAL+DCG}, The DCG mechanism enhances feature discriminability through channel-wise feature calibration, while the GAL module strengthens geometric feature representation via explicit spatial relationship modeling. The two components exhibit significant complementary advantages and jointly optimize the representation learning of sparse 3D data.

\begin{table}[h]
\centering
\resizebox{0.9\linewidth}{!}{
\begin{tabular}{c|cccc}
\hline
\multirow{2}{*}{Best results} &\multicolumn{2}{c}{S3DIS} &\multicolumn{2}{c}{MultiScan} \\
& $mAP_{25}$ & $mAP_{50}$ & $mAP_{25}$ & $mAP_{50}$\\
\hline
Only GAL & 77.8& 67.3& 67.3& 55.3\\
Only DCG  & 78.1& 67.2& 68.0& 54.1\\
GAL+DCG (Ours) &\textbf{80.5} &\textbf{71.8} &\textbf{69.6} &\textbf{56.3}\\
\hline
\end{tabular}}
\vspace{-10pt}
 \caption{Ablation study of dynamic channel gating  mechanism and geometry-aware learning.}
\label{tab:GAL+DCG}
\vspace{-15pt}
\end{table}

\begin{table}[h]
\centering
\resizebox{0.9\linewidth}{!}{
\begin{tabular}{c|cccc}
\hline
\multirow{2}{*}{Best results}&\multicolumn{2}{c}{S3DIS} &\multicolumn{2}{c}{MultiScan} \\
    & $mAP_{25}$ & $mAP_{50}$ & $mAP_{25}$ & $mAP_{50}$ \\
\hline
Manhattan & 76.8 & 67.0 & 69.1 & 56.1\\
Mahalanobis & 76.3& 65.6& 68.5 & 56.3 \\
Euclidean(Ours) &\textbf{80.5} &\textbf{71.8} &\textbf{69.6} &\textbf{56.3} \\
\hline
\end{tabular}}
\vspace{-10pt}
 \caption{Ablation study of different distance algorithms.}
\label{tab:distance algorithms}
\vspace{-10pt}
\end{table}

% To verify the effectiveness of the distance metric algorithm in the geometric perception learning module, this paper conducts an ablation study on feature centroid distance computation methods. We systematically compare three typical distance metrics: Euclidean distance, Mahalanobis distance, and Manhattan distance. As demonstrated in the table, Euclidean distance consistently outperforms both Mahalanobis and Manhattan distances. This superiority stems from Euclidean distance's better adaptability to spatial relationships. While Mahalanobis distance can capture feature distribution correlations, its performance is limited by the stability of covariance estimation. Conversely, Manhattan distance suffers from excessive sparsity constraints, leading to the loss of geometric features.

We conduct an ablation study on the feature centroid distance measurement method in the GAL module to evaluate the effectiveness of different distance algorithms. The results are shown in Table~\ref{tab:distance algorithms}, we systematically compare three typical metrics: Euclidean distance, Mahalanobis distance, and Manhattan distance. Extensive experimental results show that Euclidean distance consistently outperforms both Mahalanobis and Manhattan distances, which can be attributed to its superior adaptability to spatial relationships. Although Mahalanobis distance can capture correlations within feature distributions, its performance is limited by the stability of covariance matrix estimation. Meanwhile, Manhattan distance suffers from excessive sparsity, which inevitably leads to incomplete preservation of geometric features.

To identify the optimal value of the decay hyperparameter $\alpha$ in the GAL module, we evaluate five different configurations: $\alpha \in \left[1.0, 1.5, 2.0, 2.5, 3.0\right]$. As shown in Table~\ref{tab:alph}, the model obtains the optimum performance when $\alpha = 2.0$, demonstrating that this value strikes an optimal balance between keeping local feature sensitivity and decreasing long-range noise interference. A smaller $\alpha$ leads to insufficient decay, hence causing interference from distant places, while a larger $\alpha$ results in over-decay and loss of essential spatial contextual information.

% To identify the optimal value of classification loss weight hyperparameter $\beta$. 
% we evaluate five configurations:
We conduct an ablation study on the classification loss weight hyperparameter $\beta$,
evaluating five configurations 
% $\beta \in \left[0.3, 0.4, 0.5, 0.6, 0.7\right]$.
% We evaluate five configurations
with $\beta$ values ranging from 0.3 to 0.7.
As shown in Table~\ref{tab:beta}, the model achieves optimal performance when $\beta=0.5$. Both higher and lower $\beta$ values degrade detection accuracy. Specifically, an excessively large $\beta$ causes the model to over emphasize the classification task, resulting in insufficient optimization of localization tasks such as bounding box regression. Conversely, an excessively small $\beta$ weakens the model's ability to distinguish critical semantic features, leading to increased incorrect Predictions. Both scenarios ultimately impair detection performance.

\begin{table}[t]
\centering
\resizebox{0.9\linewidth}{!}{
\begin{tabular}{c|cccc}
\hline
\multirow{2}{*}{Best results} &\multicolumn{2}{c}{S3DIS}&\multicolumn{2}{c}{MultiScan} \\
 &$mAP_{25}$ &$mAP_{50}$&$mAP_{25}$ &$mAP_{50}$ \\
\hline
$\alpha=1.0$ &80.4 &70.8&68.0 &55.8 \\
$\alpha=1.5$ &80.0 &70.4 &68.4 &56.5\\
$\alpha=2.5$ &79.1 &69.1 &66.0 &54.6 \\
$\alpha=3.0$ &76.0 &65.8 &69.4 &\textbf{57.2}\\
$\alpha=2.0$ (Ours) &\textbf{80.5} &\textbf{71.8}&\textbf{69.6} &56.3 \\
\hline
\end{tabular}}
\vspace{-10pt}
 \caption{Ablation study of the hyperparameter $\alpha$.}
% \label{tab:}
\label{tab:alph}
\vspace{-20pt}
\end{table}
\begin{table}[h]
\centering
\resizebox{0.9\linewidth}{!}{
\begin{tabular}{c|cccc}
\hline
\multirow{2}{*}{Best results} &\multicolumn{2}{c}{S3DIS}&\multicolumn{2}{c}{MultiScan} \\
 &$mAP_{25}$ &$mAP_{50}$&$mAP_{25}$ &$mAP_{50}$ \\
\hline
$\beta=0.3$& \textbf{80.6} & 70.6 &67.5 & 56.1 \\
$\beta=0.4$ & 80.1 & 68.4 & 68.1 &56.2 \\
$\beta=0.6$ & 75.1 & 66.3 & 68.1 &55.6  \\
$\beta=0.7$ & 78.1 & 66.9 & 65.3 &52.9\\
$\beta=0.5$ (Ours) &80.5 &\textbf{71.8}&\textbf{69.6} &\textbf{56.3} \\
\hline
\end{tabular}}
\vspace{-10pt}
 \caption{Ablation study of the hyperparameter $\beta$.}
\label{tab:beta}
\vspace{-20pt}
\end{table}

% \begin{table}[h]
% \centering
% \resizebox{1\linewidth}{!}{
% \begin{tabular}{c|cccc}
% \hline
% \multirow{2}{*}{Best results} &\multicolumn{2}{c}{S3DIS} &\multicolumn{2}{c}{MultiScan} \\
% & $mAP_{25}$ & $mAP_{50}$ & $mAP_{25}$ & $mAP_{50}$ \\
% \hline
% Gaussian Function & 78.7 & 68.5& 69.3 & \textbf{58.1}\\
% Inverse Distance Function &74.5  & 66.0  & 68.0 & 56.3 \\
% Exponential Function(Ours) &\textbf{80.5} &\textbf{71.8} &\textbf{69.6} &
% 56.3 \\\hline
% \end{tabular}}
%  % \caption{Ablation study of}
% % \label{tab:}
% \end{table}

\section{Conclusion}
 In this paper, we propose UniGeo, a unified 3D indoor detection framework that integrates geometry-aware learning with a dynamic channel gating mechanism. To address the limitations of sparse 3D U-Net networks, we employ a multi-scale feature extraction strategy to capture both local details and global structural information. Moreover, we model geometric relationships through a geometry-aware learning module and enhance local feature representation via dynamic channel gating, effectively integrating global geometric information with discriminative local features. Extensive experiments on multiple indoor scene datasets demonstrate that our approach achieves state-of-the-art performance and outperforms existing mainstream methods. Ablation studies further validate the effectiveness of UniGeo.

\vfill\pagebreak

 \section{ACKNOWLEDGEMENTS}
 This work is supported by National Natural Science Foundation of China (Grant No. 62302145, 62272144), Fundamental Research Funds for the Central Universities (Grant No. JZ2025HGTB0225, JZ2024HGTG0309, JZ2024AHST0337), Major Scientific and Technological Project of Anhui Provincial Science and Technology Innovation Platform (Grant No. 202305a12020012, 202423k09020001), 
 National Key R\&D Program of China (NO.2024YFB3311602), and the Anhui Provincial Natural Science Foundation (2408085J040).

% \section{REFERENCES}
% \label{sec:refs}

% List and number all bibliographical references at the end of the
% paper. The references can be numbered in alphabetic order or in
% order of appearance in the document. When referring to them in
% the text, type the corresponding reference number in square
% brackets as shown at the end of this sentence \cite{C2}. An
% additional final page (the fifth page, in most cases) is
% allowed, but must contain only references to the prior
% literature.

% Please follow the IEEE Citation Guidelines, \url{https://ieee-dataport.org/sites/default/files/analysis/27/IEEE\%20Citation\%20Guidelines.pdf} for formatting of references.

% References should be produced using the bibtex program from suitable
% BiBTeX files (here: strings, refs, manuals). The IEEEbib.bst bibliography
% style file from IEEE produces unsorted bibliography list.
% -------------------------------------------------------------------------

\bibliographystyle{IEEEbib}
\bibliography{strings,refs}

\end{document}